\documentclass[preprint,12pt]{elsarticle}

\usepackage{amssymb}
\usepackage{amsmath}
\usepackage{multirow}
\usepackage{float}
\usepackage{longtable}
\usepackage{color}
\usepackage{lineno}
\usepackage{hyperref}

\journal{Journal of NEUROCOMPUTING}

\begin{document}

\begin{frontmatter}



\title{One for All: One-stage Referring Expression Comprehension with Dynamic Reasoning}


\author[label1,label2]{Zhipeng Zhang}

\author[label1,label2]{Zhimin Wei}

\author[label1]{Zhongzhen Huang}

\author[label1]{Rui Niu}

\author[label1,label2]{\\Peng Wang\corref{cor1}}
\cortext[cor1]{Corresponding author}

\address[label1]{School of Computer Science, Northwestern Polytechnical University, Xi'an, China}
\address[label2]{National Engineering Laboratory for Integrated Aero-Space-Ground-Ocean Big Data Application Technology, China}

\renewcommand{\baselinestretch}{0.95}

\begin{abstract}
Referring Expression Comprehension (REC) is one of the most important tasks in visual reasoning that requires a model to detect the target object referred by a natural language expression.
Among the proposed pipelines, the one-stage Referring Expression Comprehension (OSREC) has become the dominant trend since it merges the region proposal and selection stages. Many \textit{state-of-the-art} OSREC models adopt a multi-hop reasoning strategy because a sequence of objects is frequently mentioned in a single expression which needs multi-hop reasoning to analyze the semantic relation. However, one unsolved issue of these models is that the number of reasoning steps needs to be pre-defined and fixed before inference, ignoring the varying complexity of expressions. In this paper, we propose a \textit{Dynamic Multi-step Reasoning Network}, which allows the reasoning steps to be dynamically adjusted based on the reasoning state and expression complexity. Specifically, we adopt a Transformer module to memorize \& process the reasoning state and a Reinforcement Learning strategy to dynamically infer the reasoning steps. The work achieves the state-of-the-art performance or significant improvements on several REC datasets, ranging from RefCOCO (+, g) with short expressions, to Ref-Reasoning, a dataset with long and complex compositional expressions.
\end{abstract}


\end{frontmatter}


\section{Introduction}
Referring Expression Comprehension (REC) aims to detect a correct target region in an image described by a natural language expression. Therefore, we argue that the key to this task is to understand the expression and image jointly and correctly and aggregate the extracted relevant features appropriately. Only with the correct generation of the best bounding boxes in the task of the referring expression comprehension, will it benefit the downstream tasks like Visual Question Answering \cite{zhu2016visual7w,gan2017vqs,li2018tell,zhu2020simple, LAO2021541, HONG2020366} and Vision-Language Navigation \cite{wang2018look,ma2019regretful,ma2019self,tan2019learning}. \par
There are mainly two threads of work in Referring Expression Comprehension:  two-stage approaches \cite{wang2016learning,wang2018learning,plummer2018conditional,chen2017msrc,yu2017joint,yu2018mattnet} and one-stage approaches \cite{yang2019fast,yang2020improving,chen2018real,sadhu2019zero,liao2020real}. The two-stage methods first generate a series of candidate bounding boxes and then gradually select the best. While the one-stage methods attempt to combine the region proposal and selection stage. The one-stage method is not only simple and effective but also achieves great performance and therefore has become a major approach for the REC task.

\begin{figure}[H] 
\centering
\scalebox{0.57}{
\includegraphics[width=\linewidth]{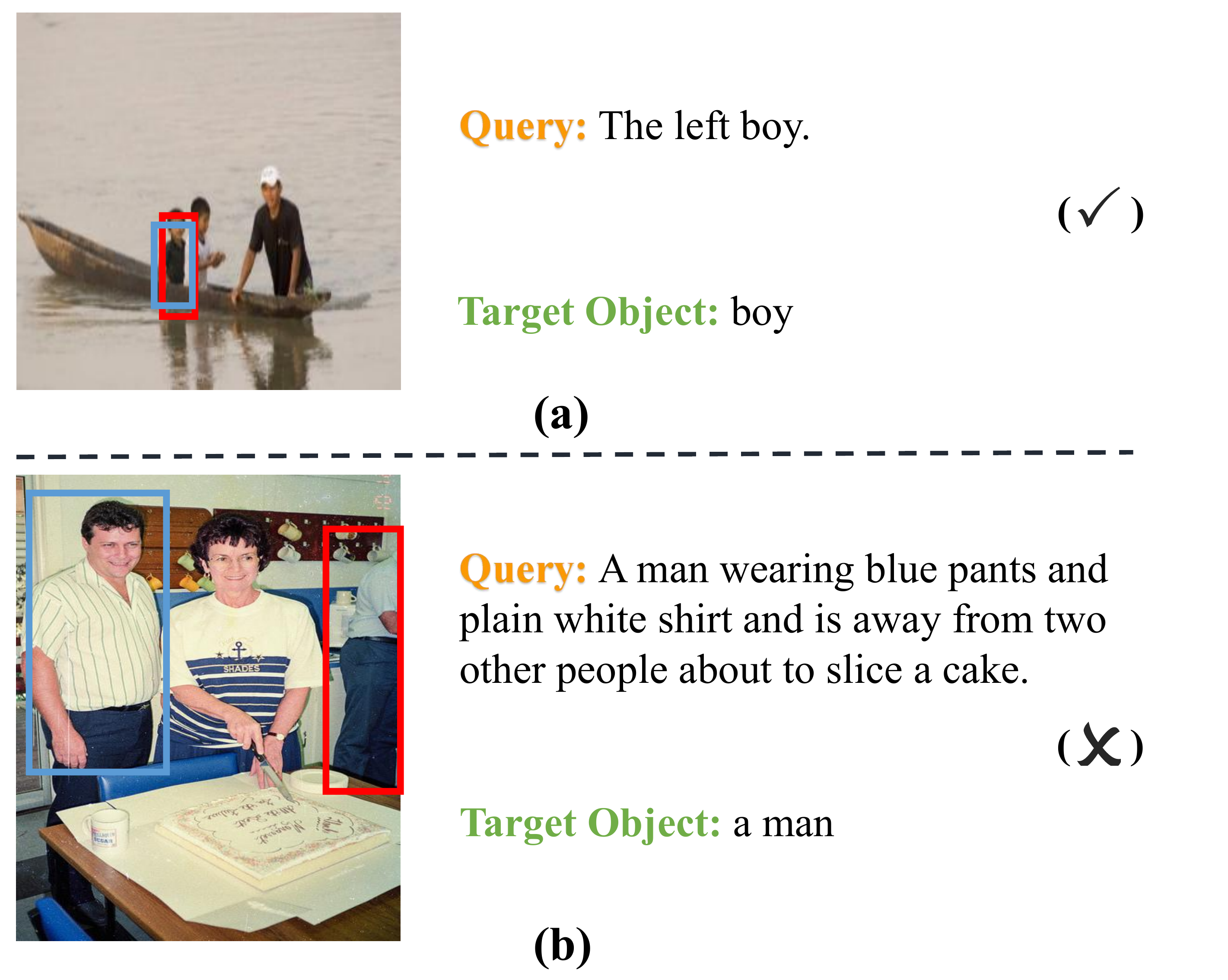}
}
\vspace{-4pt}
\caption{The red and blue boxes represent the ground-truths and predicted regions by the current \textit{state-of-the-art} one-stage method respectively. This task requires the model to ground the ground-truth region (red box) by using the input query. To get the ground-truth bounding box, the existing method needs multi-steps to reason. Existing current methods, such as \textit{ReSC}\cite{yang2020improving}, with fixed reasoning steps will search the wrong object if the expression is long and complex. For example, \textbf{ (a).} The expression is short, so we only need a few steps for reasoning the correct object. \textbf{ (b).} When facing the longer and more difficult expression, it needs more steps to reason the target object so existing methods get the wrong bounding box.}
\label{fig:introduction}
\end{figure}

However, there are also obvious flaws and shortcomings in the existing one-stage referring expression comprehension (OSREC) approaches~\cite{yang2019fast,liao2020real}. 
Most of the data processing way is encoding the expression and image as two embedding vectors and then fusing these features via concatenation for further bounding box prediction. A series of OSREC models~\cite{hu2017learning,perez2018film,yang2020graph,yang2020improving,wu2021} adopt a multi-hop fusion strategy, to reason over multiple objects mentioned in a complex expression and their spatial relations. 
Nevertheless, one unsolved issue of these models is that the number of reasoning steps is pre-defined and fixed during inference, without considering the varying complexities of different expressions. 
As is shown in Figure~\ref{fig:introduction}, 
with the increase in expression complexity, it accordingly needs more reasoning steps to reach the correct object. In this case, the \textbf{\textit{Dynamic Multi-step Reasoning Network}} proposed by us can solve the above problems primely by automatically determining the number of reasoning steps.

As \cite{si2022inception,touvron2021training,liu2022convnet}, CNNs are the most fundamental backbone for general computer vision tasks, which cover more local information through local convolution within the receptive fields, so as to effectively extract high-frequency representations. Transformer has strong capability of building long-range dependencies and fusing associations, but powerless in capturing high-frequency local information.  There are a lot of researches \cite{xu2020layoutlmv2,hu2020iterative,lin2021vx2text}  in the field of multimodal on how to combine them. Therefore, we follow their strengths to design the model.

In order to solve dynamic reasoning problems, we use reinforcement learning \cite{silver2014deterministic,khadka2018evolution,das2017learning, FAN2021422} to monitor the model and make decisions about whether to continue reasoning. Different from the previous method of fixed reasoning times, our dynamic reasoning method can automatically determine the reasoning steps according to the characteristics of the data. As a result, our model can solve the problem that the previous model faced with complex query and image relationship reasoning times are not enough, leading to prediction errors.


Our dynamic reasoning framework, \textbf{\textit{Dynamic Multi-step Reasoning Network}}, is composed of \textit{Data encoder}, \textit{Fusion Module} and \textit{Dynamic reward Module}, as \textbf{Figure~\ref{fig:arch1}} shows. Our framework dynamically chooses to continue to reason or not, according to the reasoning state and expression complexity. By introducing reinforcement learning, the recursive reasoning will be adjusted to judge automatically by our \textit{Dynamic Reward Module}. The information of expression and image will be preferably fused by our \textit{Fusion Module} regardless of the length of the expression. We use transformer to associate all features globally, to pay more attention to each detail in the image and expression.     

The main contributions of our work are as follows: 
\leavevmode \par(1). We propose a \textbf{\textit{Dynamic Multi-step Reasoning Network}} to solve referring expression comprehension tasks of diverse complexities. By adaptively selecting the number of reasoning steps during inference, our model is capable of dealing with varying-complexity expressions using the same set of hyper-parameters without manually tuning.
\leavevmode \par(2). With no need of manually tuning the reasoning steps, our model achieves the \textit{state-of-the-art} (\textit{SOTA}) performances or significant improvements on a number of REC datasets with different complexities, including RefCOCO+, RefCOCOg and Ref-Reasoning.

\section{Related Works}
\label{rw}
\subsection{Referring Expression Comprehension}
Referring Expression Comprehension (REC) is a visual-linguistic cross-modal understanding problem. It aims to detect the target object described by a natural language expression in an image. Most previous methods are two-stage methods \cite{wang2016learning,wang2018learning,plummer2018conditional,chen2017msrc,yu2017joint,yu2018mattnet}, which generate several region proposals in the first stage. The second stage is to retrieve the objected region matched with the input expression by calculating the similarity. Inspired by \textit{attention mechanism}, A-ATT \cite{deng2018visual} reasons between information jointly and further considers the self-attention guidance to explore a more diversified interaction among multiple information sources. A graph-based language-guided attention network was proposed by \textit{Wang} \cite{wang2019neighbourhood} to highlight the inter-object and intra-object relationships that are closely associated with the expression for better performance. \textit{Niu} \cite{niu2019variational} developed a variational Bayesian framework to exploit the reciprocity between the referent and context. Owing to the inaccuracy of region proposals and slow computation speed, more and more work concentrate on the one-Stage method~\cite{yang2019fast,yang2020improving,chen2018real,sadhu2019zero,liao2020real}. Compared to the two-stage method, one-stage methods directly predict bounding boxes of the object by densely fusing the features of visual-text at all spatial locations.

Recently, many works have made some advancements in the one-stage method. In the work \cite{yang2019fast}, the author puts forward a one-stage model that fuses an expression’s embedding into YOLOv3 object detector augmented by spatial features, and then it uses the merged features to localize the corresponding region. \cite{liao2020real} reformulates the REC as a correlation filtering process and puts forward CenterNet \cite{zheng2019reasoning}. The expression is first mapped from the language domain to the visual domain, and then it is treated as a template (kernel) to perform correlation filtering on the image feature maps. Besides, the peak value in the heatmap is used to predict the center of the target box. Yang \cite{yang2020improving}, uses a recursive \textit{sub-query learner} to enhance the model's ability to integrate text-image features through more steps. Although the one-stage method is effective, it also results in the loss of text features when facing a long and complicated expression. These studies have found that it needs different times instead of more times for a better result when dealing with different lengths of expressions. Hence, we propose a \textbf{\textit{Dynamic Multi-step Reasoning Network}} to reason dynamically and customize the most appropriate number of iterations for reasoning the final result. In addition, the previous attention mechanism is local, which is hard to pay attention to all detail. To solve this problem, we use \textit{transformer} to fuse features and propose \textit{Attention Module} for variable-length queries.

\subsection{Dynamic Reasoning}
Dynamic reasoning has been proposed to solve reasoning tasks. Neural Module Networks (NMNs) \cite{hu2017learning} are multi-step models that build question-specific layouts and execute. In work \cite{perez2018film}, the author raised FiLM, a multi-step reasoning procedure that influences neural network computation via a simple and feature-wise affine transformation based on conditioning information. \textit{Hu} \cite{hu2019language} proposed Language Conditioned Graph Network (LCGN) model that dynamically determines which objects to collect information from each round by weighting the edges in the graph. At the same time, since reinforcement learning is concerned with how intelligent agents ought to take actions in an environment to maximize the notion of cumulative reward, some works applied it to various vision tasks and obtained remarkable success. Recently, \textit{He} \cite{he2019read} has extended reinforcement learning to the reasoning task that formulates the task of video grounding as a problem of sequential decision making by learning an agent and improving by multi-task learning.

In recent years, more and more one-stage Referring Expression Comprehension (OSREC) have been proposed to tackle the Referring Expression Comprehension (REC) task. Moreover, in consideration of previous one-stage methods’ limitations on long and complex expressions, many OSREC models adopt a multi-hop reasoning strategy because a sequence of objects is frequently mentioned in a single long and complex expression. The work in \cite{yang2020graph} argues to learn the representations from expression and image regions in a progressive manner and performs multi-step reasoning for better matching performance. Inspired by the strategy, \textit{Luo} proposed a Multi-hop FiLM \cite{perez2018film} model that recursively reduces the referring ambiguity with different constructed sub-queries to perform multi-step reasoning between the image and language information. However, there is an unsolved issue of these models that the number of reasoning steps needs to be pre-defined, and the results appear to drop significantly when these models process long and complex expressions. Hence, we propose dynamically reason, which allows the reasoning steps to be dynamically adjusted based on the reasoning state and expression complexity for deducing the final result.

\section{The approach}
\label{the_approach}
In this section, we mainly introduce our one-stage \textbf{\textit{Dynamic Multi-step Reasoning Network}} for referring expression task. Given a natural language expression $\{q\}_{n=1}^N$, which $q$ is the word of the expression and $N$ is the length of the referring expression, the model is required to detect the described object $O$ in the given image $I$. Previous one-stage reasoning models~\cite{hu2017learning,perez2018film,yang2020graph,yang2020improving,wu2021} with multi-hop reasoning strategies either need to set the same reasoning steps for different instances or have to set different steps for various datasets through several advanced experiments. Nevertheless, detecting the correct objects from various referring expressions with different complexities needs different numbers of reasoning iterations. In addition, it's unrealizable to frequently adjust reasoning iterations and other hyper-parameters in practical use. However, the reasoning iteration steps of our model can change dynamically when the input expressions change. 

\begin{figure*}
\centering
\scalebox{1.0}{
\includegraphics[width=\linewidth]{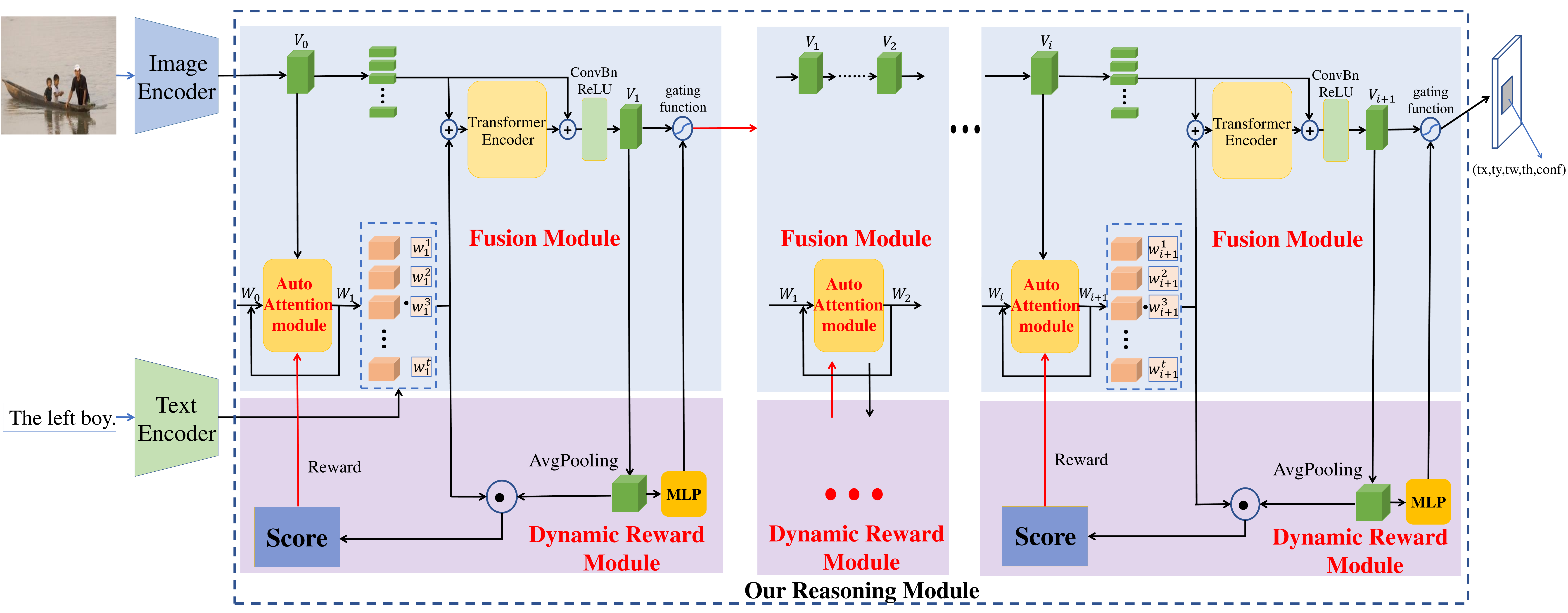}
}
\vspace{-4pt}
\caption{The overall architecture of our model, which contains three main parts: the \textit{encoder module}, the \textit{fusion module} and the \textit{dynamic reward module}. The gating function decides to continue reasoning or output the vision vector, which is activated by MLP  in \textit{Dynamic Reward Module}. There are seriatim elaborate descriptions of the three modules in Section~\ref{the_approach}. Additionally, the sign $+$ in our model means concatenating two vectors, and $\cdot$ means dot-product.
}
\label{fig:arch1}
\end{figure*}

The overall architecture of our \textbf{\textit{Dynamic Multi-step Reasoning Network}} is exhibited in {Figure~\ref{fig:arch1}}. The model can be generally divided into three components:
 (1) \textit{Data Encoder}: CNN models and BERT are adopted to extract image and expression features, respectively. (2) \textit{Fusion Module}: this module uses \textit{Transformer Encoder} to fuse visual and text features better and learns richly contextual vision-text representations. (3) \textit{Dynamic Reward Module}: a reinforcement learning strategy determining the max reasoning iterations and outputting the detected object $O$ in image $I$.

\subsection{Encoder Module}
Given an image $I\in R^{W\times H\times 3}$ and a referring expression $Q=\{q_n\}^N_{n=1}$,  where $q_n$ represents the $n$-th word and $W\times H\times 3$ denotes the spatial scale of the image, our goal is to find one sub-region $I_S$ within the image $I$ that corresponds to the semantic meaning of the referring expression $Q$.

\subsubsection{Image Encoder}
Following the work~\cite{yang2020improving}, our image encoder uses Darknet-53 pertained on COCO object detection dataset as our module's backbone to extract visual features of input images. In the experiment, We use the same initial parameters with \textit{yang} \cite{deng2018visual,yang2019fast} for fairness. Specifically, the input image was fed into the visual encoder network after being resized to the size of $256\times 256\times 3$. We select the encoded visual feature from the convolution layer with a dimension $256\times 256\times 3$, and then add $1\times 1$ convolution layer with batch normalization and \textit{ReLU} to map them all into the same dimension 512. The output feature map is $V=\{v_i\}^{W\times H}_{i=1}$,  where $W$ and $H$ mean the weight and height of the image respectively and $v_i$ is the feature $V$ of different regions $i$, which denotes different local regions for the input image.


\subsubsection{Expression Encoder}
In this paper, we use the pre-trained BERT model as our expression encoder and train with the same initial parameters \cite{yang2020improving}. In our module, each word in a referring expression which length is $N$ is firstly mapped to a corresponding word embedding vector with a linear function, giving us $X=\{x_n\}^N_{n=1}$. 
Then, each $x_n$ and its absolute position in the sentence $n$ are fed into a pre-trained BERT model. 
We summarize the representation of each word in the last layer as the vector of expression and each sentence also comes with special tokens such as [CLS], [SEP] and [PAD]. 
Finally, we obtain word-level features $E=\{e_n\}_{n=1}^N$, $e_n\in R^d$, which the dimension size $d=768$ and the maximum value of $N=20$. 

\subsection{Fusion Module}
The visual features of the images are extracted by DarkNet~\cite{bochkovskiy2020yolov4}. The referring expression $Q$ is firstly fed into a pre-trained BERT-Base~\cite{devlin2018bert} to obtain the original word features $\{e_n\}_{n=1}^N$ (dimension, $N\times768$) and then we feed the word features into an MLP, whose output dimension is $512$. 
    Our overall \textbf{\textit{Dynamic Multi-step Reasoning Network}} updates both visual and word features into ${V^t}$ and $\{e^t_n\}_{n=1}^N$ step by step which $t$ means the number of current iterations and the max reasoning iteration $T$ is decided by the \textit{Dynamic Reward Module}. 

By the inspiration from the previous work~\cite{yang2019fast,yang2020improving}, we also introduce the \textit{Attention module} into our model to construct a score vector $\{w_n^t\}_{n=1}^N$ for word features in each iteration. Otherwise,  the history of score vector is recorded to compute a history vector $\{h_n^t\}$ which helps to avoid ignoring some small tips in the referring expression. To be specific, the \textit{Attention module} takes the word
feature $\{e_n\}_{n=1}^N$, the visual feature ${V}^{t-1}$ in the last iteration, and the history vector
$\{h_n^t\}$ to compute the score.
\begin{equation}
{
	\begin{split}
	    &h_n^t = 1 - \mathbf{min}(\sum_{i=1}^{t-1}w_n^i, 1), \\
    	&w_n^t = \mathbf{softmax}[\mathbf{W}_1^t\mathbf{tanh}( \mathbf{W}_0^th_n^t(\overline{V}^{t-1}\cdot~e_n)+b_0^t)+b_1^t]
	\end{split}
	}
\end{equation}
where $\cdot$ is dot product. $\overline{V}^{t-1}$ is the average pooling of $V^{t-1}$. $\mathbf{W}_0^t$, $b_0^t$, $\mathbf{W}_1^t$ and $b_1^t$ are learnable parameters in the \textit{Attention module}. Given the history of previous score vectors, $h_n^t$ can represent how much attention each word in the expression has got so far. Both $h_n^t$ and $w_n^i$ are N-Dimension vectors
with values ranging from 0 to 1.

Considering that transformer \cite{vaswani2017attention} can make good use of global information, we adopt a 6-layer \textit{Transformer Encoder} which each layer has 8 heads into our \textit{Fusion Module} to update visual features according to the word features. By this connection, we find that the output image feature section will have a strong association with expression by using transformer construct. Multi-modal information is associated globally, which shows superior performance. The original word feature $\{e_n\}_{n=1}^N$ is weighted by multiplying the learnable score vector $\{w_n^t\}_{n=1}^N$: 
\begin{equation}
{
	\begin{split}
    	& \widetilde{e}_n = w_n^t * e_n, n=1, 2, 3, ... ,N. 
	\end{split}
}
\label{update_w}
\end{equation}

We resize the visual feature $V^{t-1}$ to $\widetilde{V}^{t-1}$. The size of feature $\widetilde{V}^{t-1}$ is $256\times512$.

Then the weighted word features $\{\widetilde{e_n}\}_{n=1}^N$ and the visual feature $\widetilde{V}^{t-1}$ are fed into the \textit{Transformer Encoder} \cite{su2019vl,carion2020end} to get updated and fused visual features $v^t$:
\begin{equation}
{
	\begin{split} 
    	& \widetilde{V}^t = \textbf{TransformerEncoder}([\{\widetilde{e_n}\}_{n=1}^N :\widetilde{V}^{t-1}]) \\
    	& \widetilde{V}^t = [\widetilde{V}^t:\widetilde{V}^{t-1}] \\
    	& V^t = \textbf{Resize}(\textbf{ConvBNReLU}(\widetilde{V}^t))
	\end{split}
}
\end{equation}
where $[:]$ indicates concatenation. Then the updated and fused visual features are used to calculate the weights of all words when the step is $t+1$. In the last step, the gating function decides to continue reasoning or output the vision vector, which is activated by the \textit{Dynamic Reward Module}. 
Our \textit{Fusion Module} uses the visual-text contextual feature as input to output the predicted grounding for the referring expression in the final iteration. Following the work \cite{yang2020improving}, We adopt the same two $1\times1$ convolutional layers to predict $9$ anchor boxes at each location where there are $32\times32=1024$ spatial locations. For each box, our module predicts five values ${t_x, t_y, t_w, t_h, conf}$. The last parameter is the confidence score and other parameters represent the relative offsets.

\subsection{Dynamic Reward Module}
In \textit{Dynamic Reward Module}, we employ policy gradient (PG) \cite{silver2014deterministic,khadka2018evolution,das2017learning} to set up reward and punishment mechanisms to optimize the final decision of the model. We resort to the policy gradient for outputting every action's probability value after each iteration to decide whether to continue reasoning or output the $V^t$.

The \textit{Dynamic Reward Module} comprises two kinds of actions helping to decide whether the system will continue reasoning or not. The action states are dependent on current visual feature $V^t$,  word features $\{e_n\}_{n=1}^N$ and $\{w_n^t\}_{n=1}^N$. The probability values are calculated by:

\begin{equation}
{
	\begin{split}
    	& \hat{e} =e_{cls}, \\
    	& \hat{V}^t = \mathbf{AvgPooling}(V^t), \\
    	& actions\_prob = \mathbf{softmax}[\mathbf{W_2}^t\mathbf{tanh}( \mathbf{W_1}^t[\hat{V}^t:\hat{e}]+b_1^t)+b_2^t] \\
    	& action = \mathbf{argmax}(actions\_prob) \\
	\end{split}
	}
	\label{actionk}
\end{equation}
where $argmax$ returns the indices of the maximum values in \textit{actions\_prob}. The feature of vision vector $V^t$ passes through the $\mathbf{AvgPooling}$. The parameters of the \textit{Attention module} $\mathbf{W}$ and $b$ are updated and learned under this section reward. They work to compute \textit{actions\_prob} and \textit{action}. The \textit{action} includes two kinds of signals, which $0$ represents stopping reasoning and $1$ represents continuing reasoning. The \textit{actions\_prob} represents the predicted probabilities of two actions.

We deal with the input expression at the first token's position and the last token's location ([CLS],[SEP]). We also join the [CLS] as the expression vector representation in MLP from BERT-encoder into our \textit{Dynamic Reward Module} to be better rewarded. 
The judgment of MLP works with the connected vector.

To train \textit{Dynamic Reward Module}, we set two kinds of reward: the \textit{ultimate reward} and the \textit{immediate reward} and adopt both of them. The details about them are as follows.
\subsubsection{Ultimate Reward}
The \textit{ultimate reward} represents the reward to the result of reasoning. We use the output of the last reasoning step to regress the bounding box of the target object $O$. The \textit{ultimate reward} is defined as:
\begin{equation}
{
	\begin{split}
    	& r_{ultimate}^t = \left\{\begin{array}{cc}
    	     1, & IoU >= 0.5, \\
    	     -1, & otherwise, \\
    	\end{array},
    	\right.
	\end{split}
}
\end{equation}
where IoU is calculated by comparing prediction bounding boxes with the ground truth bounding box. The \textit{ultimate reward} is 1 if the IoU is greater than 0.5. Otherwise, the \textit{ultimate reward} is -1.

\subsubsection{Immediate Reward}
The \textit{Immediate Reward} represents the reward to the positive effect of features during reasoning. The reasoning module concentrates on the most related words and visual field and makes visual features and weighted word features more and more relative step by step. Accordingly, we calculate the \textit{immediate reward} between the visual features and weighted word features during all the reasoning steps. The reward is calculated as:
\begin{equation}
{
	\begin{split}
	    & L^t = \sum_{n=1}^{N}w_n^t * e_n, \\
	    & \hat{V}^t = \mathbf{AvgPooling}(V^t), \\
	    & Score^t = L^t\cdot\hat{V}^t, \\
    	& r_{imm}^t = \left\{\begin{array}{cc}
    	     1, & Score^t-Score^{t-1} >= 0, \\
    	     -1, & otherwise, \\
    	\end{array}.
    	\right.
	\end{split}
	}
\end{equation}
Where $Score^t$ is the relevancy degree of weighted word features and visual features in the $t$-th reasoning step. The \textit{immediate reward} is 1 if the relevancy degree is increasing. Otherwise, the \textit{immediate reward} is -1.

The data results of these different reward policies are shown in Table~\ref{rewards}. The final model framework is compatible with the use of both kinds of rewards in order to achieve the best possible results. The gating function $\mathbf{r^t}$ is also computed in this step. The model will continue to reason if the gating function $\mathbf{r^t}\textgreater0$, which means that $r_{imm}^t=r_{ultimate}^t=1$.

To train the Dynamic Reasoning globally, we use $actions\_prob$ as the prediction and the $action$ is used as our label. Accordingly, we use the weighted CrossEntropyLoss between the softmax over all boxes and a one-hot vector — the anchor box, which has the highest IoU with the ground truth region, is labeled 1 and all the others are labeled 0 as the loss function.
The weight is calculated by:
\begin{equation}
{
	\begin{split}
	    & r^t = r_{ultimate}^t + r_{imm}^t \\
	    & weight^t = \sum_{t=i}^T0.9^{t-i}r^t, i = 1, 2, 3, ...,T \\
	\end{split}
	}
\end{equation}

\section{Experiment}

In this section, we conduct experiments to analyze our model. Firstly, the proposed model is compared with a variety of REC models on different datasets. Especially, we compare our model with other excellent one-stage methods to analyze the effectiveness of our reasoning architecture and reinforcement learning. Then, a series of ablation experiments are performed to analyze the effectiveness of different rewards, the impact of different iterations and the impact of different transformers.

All the experiments are conducted on $8$ Nvidia TitanX GPUs. The proposed model is implemented with PyTorch. Following the universal setting \cite{yang2020improving,yang2019fast},  we keep the original image ratio and resize the long edge to 256 and then pad the resized image to $256 \times 256$ with the mean pixel value. The RMSProp optimizer with a learning rate of $1e^{-4}$ initially which decreases by half every 10 epochs is used to train the model. We adopt a batch size of $8$ and train the model with $100$ epochs. Consistent with previous work, we also use accuracy as the evaluation metric, which is calculated by checking whether the target object is correctly selected or not. Given a language expression, the predicted region is considered as the correct grounding if the Intersection-over-Union(IoU) score with the ground-truth bounding box is greater than 0.5.

\subsection{Datasets and Analysis Benchmark}
\noindent\textbf{RefCOCO/RefCOCO+/RefCOCOg.}  RefCOCO \cite{10.1007/978-3-319-46475-6_5}, RefCOCO+ \cite{10.1007/978-3-319-46475-6_5},
and RefCOCOg \cite{mao2016generation} are three Referring Expression Comprehension datasets with images and referred objects selected from MSCOCO \cite{lin2014microsoft}. The referred objects are selected from the MSCOCO object detection annotations and divided into 80 object classes. RefCOCO has 19,994 images with 142,210 referring expressions for 50,000 object instances. RefCOCO+ has 19,992 images with 141,564 referring expressions for 49,856 object instances. RefCOCOg has 25,799 images with 95,010 referring
expressions for 49,822 object instances. It was collected in a non-interactive setting thereby producing longer expressions than that of the other three datasets which were collected in an interactive game interface. On RefCOCO and RefCOCO+, we follow the general standard split of train/validation/testA/testB that has 120,624/ 10,834/ 5,657/ 5,095 expressions for RefCOCO and 120,191/ 10,758/ 5,726/ 4,889 expressions for RefCOCO+ respectively. Images in “testA” are of multiple people
while images in “testB” contain all other objects. RefCOCO+ is similar to RefCOCO where however, absolute location words are forbidden to use , so it takes more effort obviously. To be specific, the expressions of RefCOCO+ do not contain words with positional relationship attributes. For example, “on the right” describes the object’s location in the image. On RefCOCOg, we experiment with the splits
of RefCOCOg-umd \cite{nagaraja2016modeling} and refer to the splits as the val-u and test-u in Table ~\ref{acc}. The queries in RefCOCOg are generally longer than those in RefCOCO and RefCOCO+: the average lengths are 3.61, 3.53, 8.43 for RefCOCO, RefCOCO+, RefCOCOg respectively. In fact, with the development of REC tasks, the descriptions become longer and more complex and similar to human real-world social language expressions. So our mission is of great significance.

\noindent\textbf{Ref-Reasoning.}  Ref-Reasoning \cite{yang2020graph} is built on the scenes from the GQA dataset \cite{hudson2019gqa} and includes semantically rich expressions, which describe objects, attributes, direct relations and indirect relations with different layouts. The numbers of the expression-referent pairs for training, validation and test on the dataset are 721,164, 36,183 and 34,609 respectively. The referring expressions for each image are generated based on the image scene graph using a set of templates and diverse reasoning layouts. In total, there are 1,664 object classes, 308 relation classes and 610 attribute classes in the adopted scene graphs. 

\begin{figure}[H]
\centering
\scalebox{0.48}{
\includegraphics[width=\linewidth]{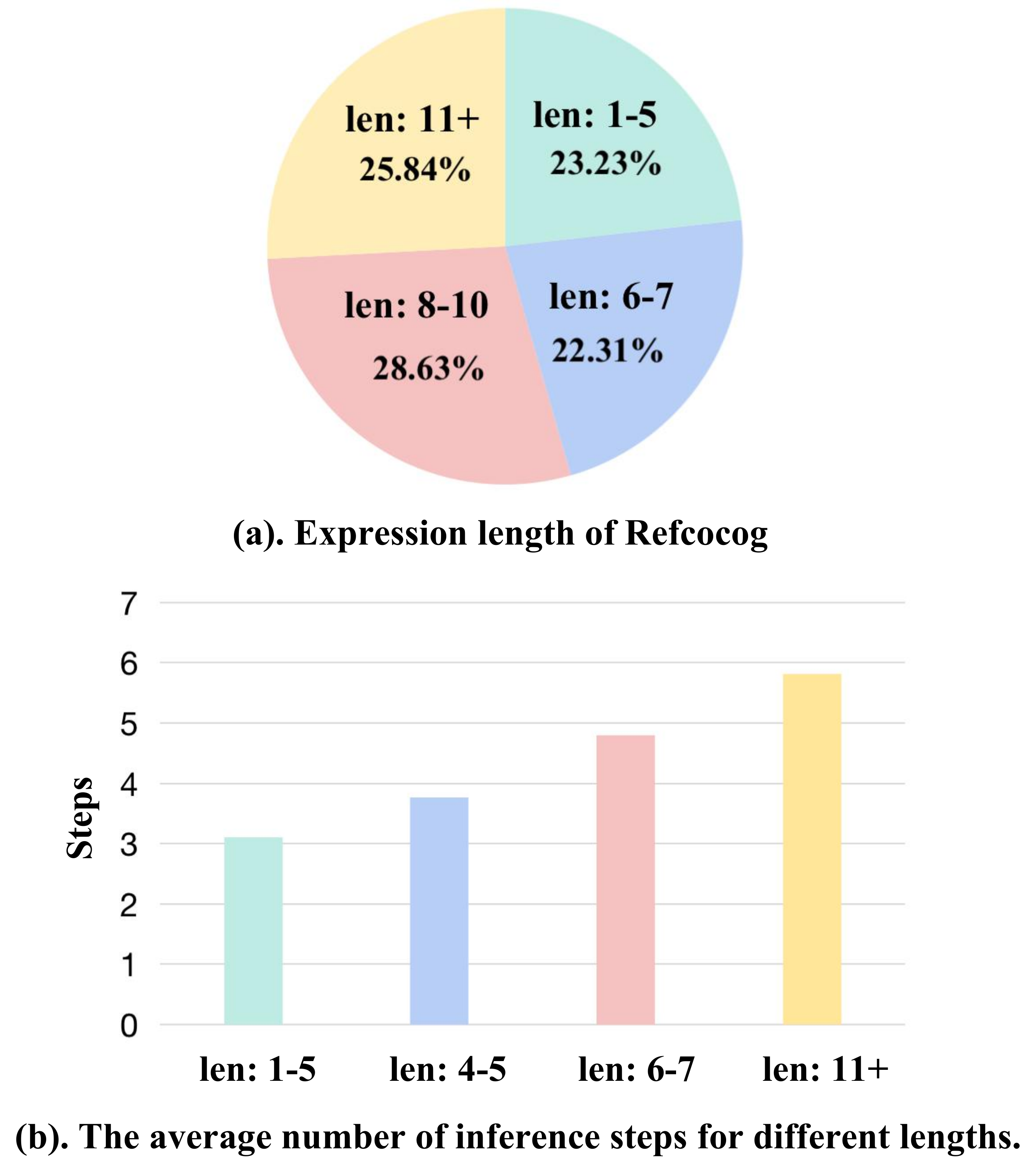}
}
\vspace{-4pt}
\caption{The proportion of different referring expressions length in RefCOCOg and the analysis of the average number of inference steps required for different lengths. }
\label{fig:relative}
\end{figure}

We present the statistical visualization results in \textbf{Figure~\ref{fig:relative}} using the common division criteria. As shown in \textbf{Figure~\ref{fig:relative}}, there is no significant distinction between the percentage of sentences of different range lengths in Refcocog.

\subsection{Comparison with State-of-the-arts}
 In the following, previous models are evaluated on different datasets and compared with our proposed model.

The overall accuracy of all evaluated models as well as the proposed model is presented in Table~\ref{acc}. In the table, we can see that compared with previous one-stage methods, our method achieves remarkable improvements. Especially, our model outperforms Sub-query-Base \cite{yang2020improving} (one-stage baseline) by nearly $2.69\%$, $2.00\%$ and $1.10\%$ on RefCOCOg-umd \cite{nagaraja2016modeling}, RefReasoning dataset \cite{yang2020graph} and RefCOCO split \cite{10.1007/978-3-319-46475-6_5} datasets. Since the expressions in RefCOCOg are generally longer, the great improvements on RefCOCOg mean that our model is very capable of handling long and difficult expressions. For fair comparisons, we don't take account of two-stage methods and Sub-query-Large \cite{yang2020improving, wang-etal-2021-fine-grained}. As \cite{yang2020improving} says, two-stage methods highly rely on the region proposals' quality. The COCO-trained detector can generate nearly perfect region proposals on COCO-series datasets for two-stage methods, but not for one-stage methods. When dealing with other datasets, their performances drop dramatically. Relatively speaking, our model is stable across all datasets. As for Sub-query-Large \cite{yang2020improving}, it has a larger image size and another BERT-large in the process of image encoder and text encoder which differ from us. Therefore, we choose Sub-query-Base as the comparison baseline of the experiments. In total, our model achieves \textit{SOTA} performances in REF-reasoning and performs better generally compared with previous one-stage methods and achieves nearly sota performances in RefCOCO(+, g) \cite{MDETR2021} . 
\begin{table*}[htb]
	\caption{Performance (Acc\%) comparison with the state-of-the-art methods and our proposed model on the RefCOCO, RefCOCO+, RefCOCOg and RefReasoning datasets especially with one-stage methods. Our proposed model achieves excellent performance among one-stage models.}
    \label{acc}
    {

    	\begin{center}
            {
                \resizebox{\textwidth}{!}{
    				\begin{tabular}{c|l|c|c|c|c|c|c|c|c|c|c}
    					\hline
    					\multicolumn{1}{c|}{\multirow{2}{*}{Type}} &  \multicolumn{1}{c|}{\multirow{2}{*}{Method}} & \multicolumn{1}{c|}{\multirow{2}{*}{Backbone}} & \multicolumn{3}{|c}{RefCOCO} & \multicolumn{3}{|c}{RefCOCO+} & \multicolumn{2}{|c}{RefCOCOg} &  \multicolumn{1}{|c}{RefReasoning}
    					\\\cline {4-12} ~ & ~ & ~ & val & testA & testB & val & testA & testB  & val-u & test-u & val \\
    					\hline \hline
    					$ $ & LGARNs \cite{wang2019neighbourhood} & VGG16 & $-$ & $76.60$ & $66.40$ & $-$ & $64.00$ & $53.40$  & $-$ & $-$ & $-$ \\
    					$ $ & MAttNet \cite{yu2018mattnet} & Res101 & $76.40$ & $80.43$ & $69.28$ & $64.93$ & $70.26$ & $56.00$ & $66.67$ & $67.01$ & $-$ \\
    					Two-stage & DGA \cite{yang2019dynamic} & Res101 & $-$ & $78.42$ & $65.53$ & $-$ & $69.07$ & $51.99$ & $-$ & $63.28$ & $-$ \\
    					Models & RvG-Tree \cite{8691415} & Res101 & $75.06$ & $78.61$ & $69.85$ & $63.51$ & $67.45$ & $56.66$  & $66.95$ & $66.51$ & $-$ \\
                        $ $ & NMTree \cite{9009000} & Res101 & $76.41$ & $81.21$ & $70.09$ & $66.46$ & $72.02$ & $7.52$  & $65.87$ & $66.44$ & $-$ \\
    					$ $ & SGMN \cite{yang2020graph} & FasterR-CNN & $-$ & $-$ & $-$ & $-$ & $-$ & $-$ & $-$ & $-$ & $25.5$ \\
    					\hline
    					$ $ & SSG\cite{chen2018real} & DarkNet53 & $-$ & $76.51$ & $67.50$ & $-$ & $62.14$ & $49.27$  & $58.80$ & $-$ & $-$ \\
    					$ $ & FAOA\cite{9010627} & DarkNet53 & $72.54$ & $74.35$ & $68.50$ & $56.81$ & $60.23$ & $49.60$  & $61.33$ & $60.36$ & $-$ \\
    					 One-Stage & One-Stage Bert \cite{yang2019fast} & DarkNet53 & $72.05$ & $74.81$ & $67.59$ & $55.72$ & $60.37$ & $48.54$  & $59.03$ & $58.70$ & $-$ \\
    					 Models & Sub-query-Base\cite{yang2020improving} & DarkNet53 & $76.74$ & $78.61$ & $71.86$ & $\textbf{63.21}$ & $65.94$ & $\textbf{56.08}$ & $64.89$ & $64.01$ & $29.5$ \\
    					
    					$ $ & \textbf{Our model} & DarkNet53 & $\textbf{76.99}$ & $\textbf{79.71}$ & $\textbf{72.67}$ & $61.58$ & $\textbf{66.60}$ & $54.00$  & $\textbf{66.03}$ & $\textbf{66.70}$  & $\textbf{31.5}$ \\
    					\hline
    				\end{tabular}}
    			}
    		\end{center}
    }
    
\end{table*}

We show the distribution of relatives between iterations of reasoning and the length of referring expression on the RefCOCOg test-umd split in \textbf{Figure~\ref{fig:relative}}. According to our statistics, we can conclude that longer and more complex expressions do require more inference steps to make the module integrate features fully. Without a suitable and sufficient number of steps, the model is not able to fuse features adequately, which will affect the subsequent decision in the candidate regions. So one of the reasons for the improvement of accuracy is that we can adapt it to automatically determine the inference step for all expression lengths.

\begin{figure}[htb]

\centering
\scalebox{0.4}{
\includegraphics[width=\linewidth]{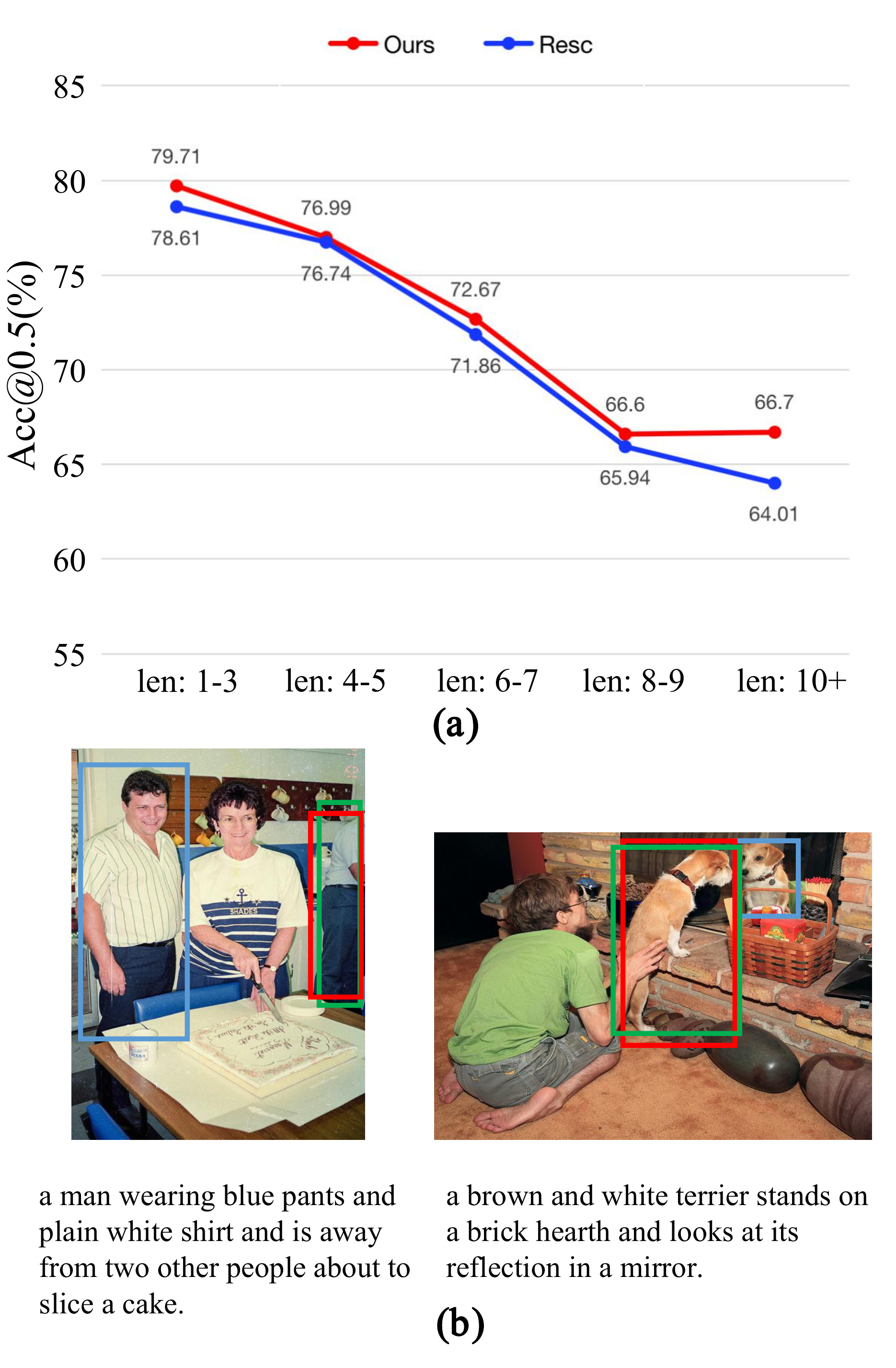}
}
\vspace{-4pt}
\caption{The comparison of accuracy on different lengths of expressions between \emph{ours} and \emph{Resc}. The red region is the true target area and our module proposal region is green. The blue circled region is detected by the previous work Resc model. We can see that the error will easily happen when facing the long and difficult expression.}
\label{fig:acc-resc}
\end{figure}

We also visualize some experimental results in \textbf{Figure~\ref{fig:acc-resc}} to show the process of reasoning confronted with the strength of expressions. The result shows, \textbf{(a).} that the advantage of our model will gradually show up as the length of the expression increases. \textbf{(b).} We visualize the two results between ours and Resc's \cite{yang2020improving}. Our \textit{Dynamic Multi-step Reasoning Network} shows the superior performance when tackling longer expression. The previous work always predicts the wrong bounding box. We believe that in the fusion between image feature and expression feature, without correct reasoning steps, the model can't choose the truth by comparing the various regions at a fixed inference step setting,  especially for long expression descriptions. 

In \ref{ablation} {Ablation Studies}, through these, the role played by each part of our model. In the following, we will analyze the role of each part in detail.

\subsection{Ablation Studies}
\label{ablation}
In this section, we elaborate on the details of the ablation studies to prove the superiority and validity of our model. First, we illustrate the impact of different iterations by setting a range of numbers of iterations to show how iterations influence the model's performance. Then, we demonstrate the effectiveness of rewards adopted in our method and the impact of the parameters of the \textit{transformer encoder} by a sequence of comparison experiments. 
\subsubsection{Impact of Different Iterations}  Experiments are conducted on RefCOCOg, which has a wider distribution of referring expressions' length as \textbf{Figure~\ref{fig:relative}} shows. We can see how the model with different iterations performs when processing datasets of various referring expression lengths.

Each row in Table~\ref{iterations} displays the result of the experiments where the numbers of iterations are set to 1, 3, 5, 8, 10 respectively. Table~\ref{iterations} shows that, with the maximal iterations increasing, the result will improve in general. But if the maximal iteration increases too much, the performance will decrease instead. It is worth mentioning that all the current works also confirm the phenomenon in our experiment, i.e., the bigger number of iterations does not mean better performance. If the number of iterations is set too much like  8, 10, etc. for better performance in long queries, it will sacrifice the reasoning time and performance in short queries. In the meanwhile, the experiment further demonstrates that our \textit{Dynamic Multi-step Reasoning Network} can make correct decisions for variable-length queries and customize the most appropriate number of iterations for deducing the final result, which has stronger robustness than the previous models with a fixed number of iterations. 
\begin{table*}[h]
	\caption{The performance (Acc\%) of the model under different iterations.}
	\label{iterations}
	{
	    \begin{center}
	    {
	    
				\begin{tabular}{l|c|c}
					\hline
					\multicolumn{1}{c|}{\multirow{2}{*}{Method}} & \multicolumn{2}{|c}{RefCOCOg}
					\\ \cline{2-3}	&  val-u & test-u \\
					\hline \hline
					our model (without RL, 1 pass) & $60.40$ & $60.73$ \\
					our model (without RL, 3 pass) & $65.81$ & $66.15$ \\
					our model (without RL, 5 pass) & $\textbf{66.01}$ & $\textbf{66.56}$ \\
					our model (without RL, 8 pass) & $65.72$ & $66.08$ \\
					our model (without RL, 10 pass) & $65.43$ & $65.87$ \\
					\hline
				\end{tabular}}
			\end{center}

	}
\end{table*}

\subsubsection{Effectiveness of Different Rewards}
To validate the effectiveness of \textit{ultimate reward} and \textit{immediate reward}, we compare it with a model without any rewards and a model without \textit{immediate reward}. As is shown in Table~\ref{rewards}, the result improves by $0.53\%$ on the test-umd split with the \textit{ultimate reward} and the result improves by $0.14\%$ with \textit{immediate reward}, which as the whole contributes to an improvement of $\textbf{0.77\%}$. It suggests that compared with \textit{ultimate reward} and \textit{immediate reward} can fulfill predominant improvement, which simultaneously indicates that our reasoning module can focus on the most related words and visual field, and make visual features and weighted expression features more and more relative step by step. 

\begin{table}[H]
	\caption{The performance (Acc\%) of the model with different rewards. The final model with both two reward mechanisms applied presents superior performance. }
	\label{rewards}
	{
	    \begin{center}
	    {
				\begin{tabular}{l|c|c}
					\hline
					\multicolumn{1}{c|}{\multirow{2}{*}{Method}} & \multicolumn{2}{|c}{RefCOCOg}
					\\ \cline{2-3}	&  val-u & test-u \\
					\hline \hline
					our model + without rewards & $65.24$ & $65.56$ \\
					our model + ultimate reward  & $66.01$ & $66.59$ \\
					our model + immediate reward & $65.81$ & $66.23$ \\
					\hline
					our final model & $\textbf{66.03}$ & $\textbf{66.70}$ \\
					\hline
				\end{tabular}
			}
			\end{center}
	}

\end{table}

\subsubsection{Impact of the Parameters of Transformer Encoder}
We also conduct experiments on the RefCOCOg dataset with different numbers of the layers and heads of Transformer \cite{vaswani2017attention} to figure out the influence of different \textit{Transformer Encoders}. Table~\ref{trans} lists out the results of different settings of the parameters, which show that the performance will improve with the increment of the layer and head. But for the maximum setting, the performance will also decrease. Additionally, by contrasting the results with the same number of heads and a different number of layers (for example, row 1 and row 2 or row 3 and row 4), we observe that the increment of layers is more crucial for the performance improvement. This experimental result clearly indicates that our proposed \textit{Transformer Encoder} is capable of extracting and fusing the information from images and expressions, which plays a vital part in the subsequent dynamic inference process. Meanwhile, \textit{Transformer Encoder} can achieve great performance for further extraction of the fused information at each iteration. 
But to some extent, the constant increment will cause performance degradation, like the result of row 5 in Table~\ref{trans}.

\begin{table}[H]
	\caption{ The performance (Acc\%) of our model under different Transformer Encoders with different parameters without containing the Dynamic Reward module.}
	\label{trans}
	{
        \begin{center}
		{
				\begin{tabular}{l|c|c}
					\hline
					\multicolumn{1}{c|}{\multirow{2}{*}{Method}} & \multicolumn{2}{|c}{RefCOCOg}
					\\ \cline{2-3}	&  val-u & test-u \\
					\hline \hline 
					our model(without RL, 1 layer 1 head) & $48.87$ & $48.93$ \\
					our model(without RL, 6 layer 1 head) & $64.09$ & $64.25$ \\
					our model(without RL, 1 layer 8 head) & $49.89$ & $49.95$ \\
					our model(without RL, 6 layer 8 head) & $\textbf{65.52}$ & $\textbf{65.70}$ \\
					our model(without RL, 8 layer 16 head) & $64.19$ & $64.61$ \\
					\hline
				\end{tabular}
			}
		\end{center}

	}

\end{table}

\subsection{Qualitative Results Analyses}
\begin{figure}[h]
\centering
\scalebox{1.0}{
\includegraphics[width=\linewidth]{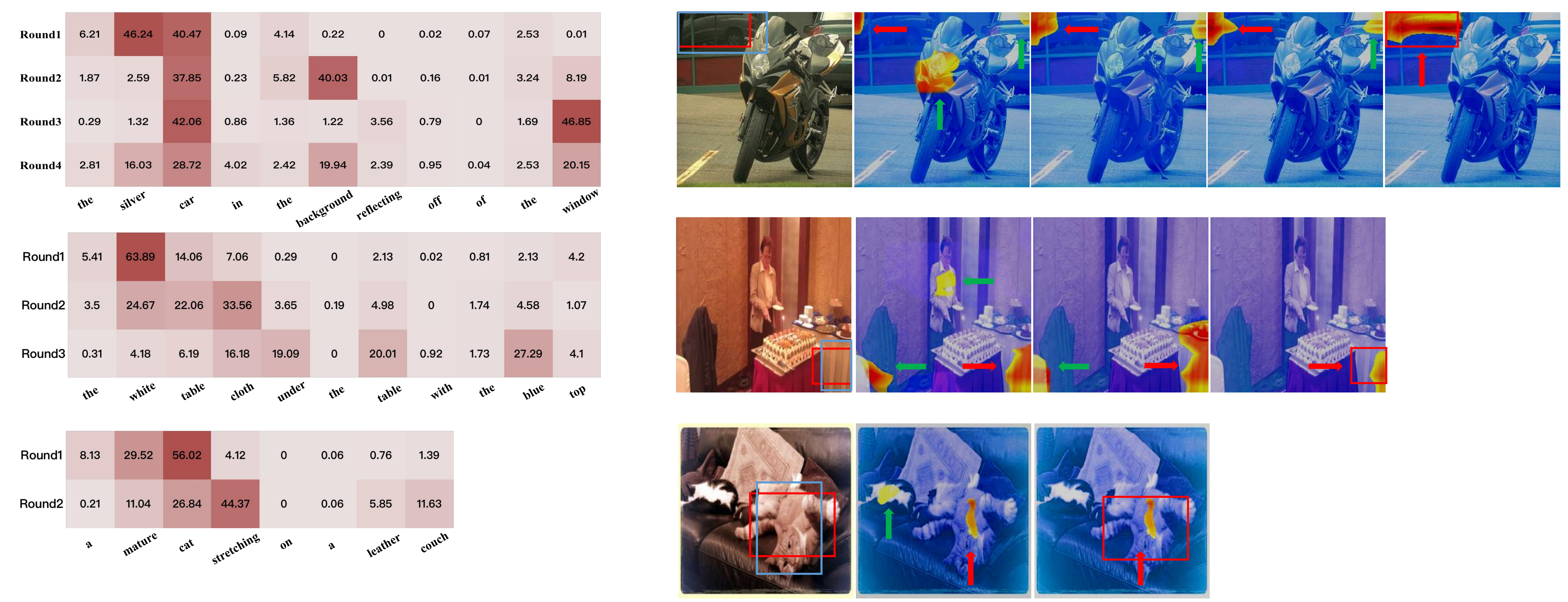}
}
\vspace{-4pt}
\caption{Visualization of one-stage referring expression comprehension with dynamic reasoning and visual feature at each step. Red/ Blue boxes are the predicted regions/ ground truth. The red arrow and the green arrow point to the target and the major distracting object on heatmaps respectively. Our model can make decisions to use different rounds for reasoning the target in the situation of various lengths of queries.
}
\label{fig:visualization}
\end{figure}

In order to analyze the advantages and weaknesses of our method, we present a visualization of the model processing as a way to explain how our model works in every step for the task in \textbf{Figure~\ref{fig:visualization}}. On the left of Figure~\ref{fig:visualization}, we can see that the performance of our BERT-Based attention mechanism expression encoder vector is enhanced gradually in the processing of representations. In addition, we can see that our attention region is getting smaller and more precise with multiple rounds of inference on the right. It will be very important for downstream tasks like robotics, UAV vision language navigation \cite{wang2018look,ma2019regretful,ma2019self,tan2019learning}, etc.

\begin{figure}[H]
\centering
\scalebox{0.9}{
\includegraphics[width=\linewidth]{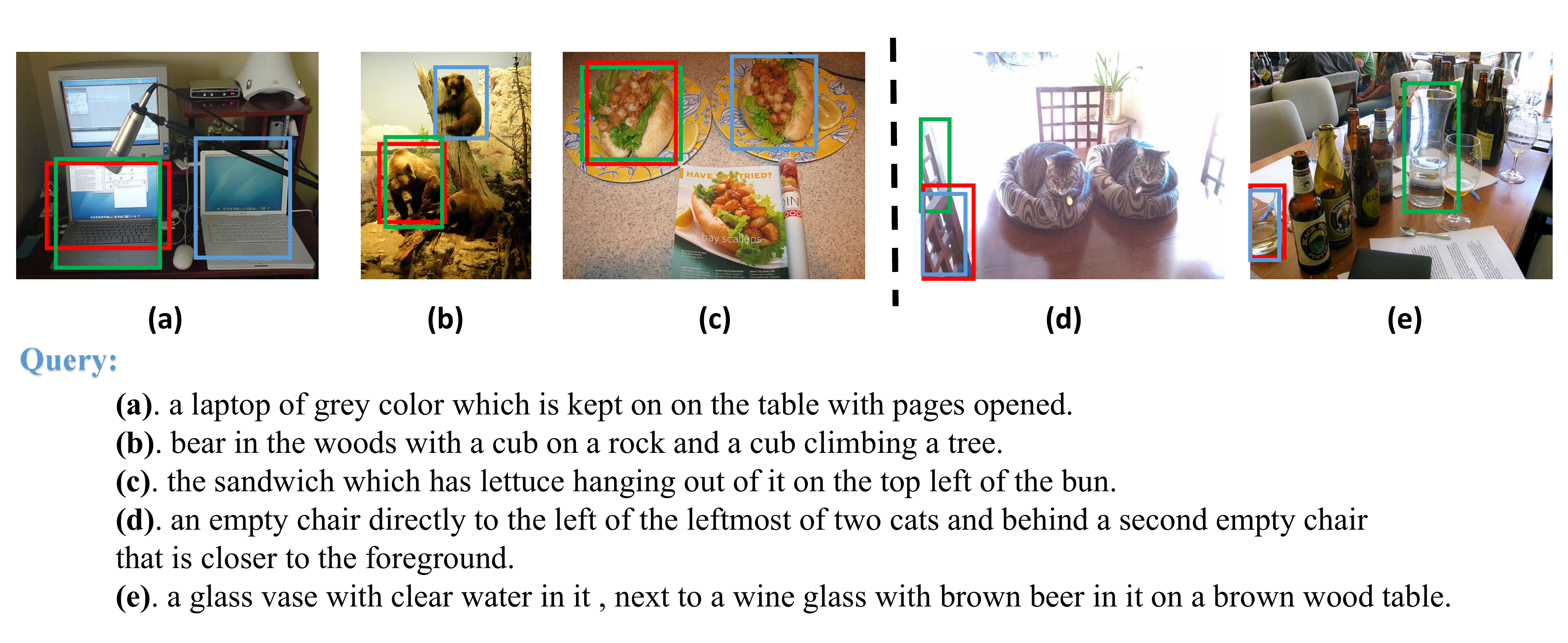}
}
\vspace{-4pt}
\caption{ The successful and failed cases of our method and the comparison between the performance of our method and the current \textit{state-of-the-art} one-stage method in handling long and complex queries. Green/blue boxes are ground-truths/predicted regions by the current \textit{state-of-the-art} one-stage method, and the red ones represent predicted boxes by our model. The three pictures on the left are successful, and the right pictures are some failures in our method.
}
\label{fig:comparasionlast}
\end{figure}

Finally, Figure~\ref{fig:comparasionlast} shows the successful and failed cases of our method and comparison. Our method performs better when the expressions are long and complicated in most instances. As is shown in (a) (b) (c), the effect of our method  is obviously improved. The right pictures (d) and (e) are two failure cases of our method. In (d), the error occurs because the multiple relations of the objects are blurred and the spatial position information is adjacent. In (e),  the detection fails to result from the ambiguity of the objects' description.

\section{Conclusion}
In this work, we propose a \textit{Dynamic Multi-step Reasoning Network}, which solves the issue that exists in one-stage methods, the unsure numbers of reasoning steps. We use the same hyper-parameters for all of the above Datasets tasks without tuning the number of reasoning steps for each. Experiments show that with the effective integration of Transformer module and Reinforcement Learning strategy, our model can automatically determine the number of inferences and achieve the \textit{state-of-the-art} performance or significant improvement on several REC datasets universally in spite of the length and complexity of expression. We hope this method will enlighten the community to move forward with the research of one-stage referring expression comprehension.




\end{document}